\documentclass[10pt,twocolumn,letterpaper]{article}

\usepackage{cvpr}
\usepackage{times}
\usepackage{epsfig}
\usepackage{graphicx}
\usepackage{amsmath}
\usepackage{amssymb}
\usepackage{color}
\usepackage{booktabs}

\usepackage{subcaption}

\providecommand\KGtwo[1]{\textcolor{black}{#1}}
\providecommand\cc[1]{\textcolor{black}{#1}}
\providecommand\KG[1]{\textcolor{black}{#1}}
\providecommand\BX[1]{\textcolor{black}{#1}}
\providecommand\bx[1]{\textcolor{black}{#1}}
\providecommand\YK[1]{\textcolor{black}{#1}}
\providecommand\RV[1]{\textcolor{black}{#1}}

\newcommand\blfootnote[1]{%
  \begingroup
  \renewcommand\thefootnote{}\footnote{#1}%
  \addtocounter{footnote}{-1}%
  \endgroup
}

\usepackage[breaklinks=true,bookmarks=false]{hyperref}

\cvprfinalcopy % *** Uncomment this line for the final submission

\def\cvprPaperID{762} % *** Enter the CVPR Paper ID here

\begin{document}

%%%%%%%%% TITLE
%\KG{Short and sweet?  Less is more?} The shorter, the better:
\title{Less is More: Learning Highlight Detection from Video Duration}

\author{Bo Xiong$^{1}$, Yannis Kalantidis$^{2}$, Deepti Ghadiyaram$^{2}$, Kristen Grauman$^{3*}$
\\
$^{1}$ University of Texas at Austin, $^{2}$ Facebook AI, $^{3}$ Facebook AI Research\\
{\tt\small bxiong@cs.utexas.edu,\{yannisk,deeptigp,grauman\}@fb.com}}

\maketitle
%\thispagestyle{empty}

%%%%%%%%% ABSTRACT
\begin{abstract}
\blfootnote{$^{*}$ \emph{On leave from University of Texas at Austin (grauman@cs.utexas.edu).}}

  Highlight detection has the potential to significantly ease video browsing, but existing methods often suffer from expensive supervision requirements, where human viewers must manually identify highlights in training videos. We propose a scalable unsupervised solution that exploits \emph{video duration} as an implicit supervision signal. Our key insight is that video segments from shorter user-generated videos are more likely to be highlights than those from longer videos, since users tend to be more selective about the content when capturing shorter videos.  Leveraging this insight,  we introduce a novel ranking framework that prefers segments from shorter videos, while properly accounting for the inherent noise in the (unlabeled) training data. We use it to train a highlight detector with 10M hashtagged Instagram videos.  In experiments on two challenging public video highlight detection benchmarks, our method substantially improves the state-of-the-art for unsupervised highlight detection.
\end{abstract}

\section{Introduction}\label{sec:introduction}

\KG{\emph{``I didn't have time to write a short letter, so I wrote a long one instead.''} -- Mark Twain}
\vspace*{0.05in}

The video overload problem  is intensifying.  With the increasing prevalence of portable computing devices (like smartphones, wearables, etc.) and promotion from social media platforms (like Facebook, Instagram, YouTube), it is seamless for Internet users to record and share massive amounts of video. According to Cisco~\cite{cisco}, by 2021 video traffic will be 82\% of all consumer Internet traffic, and every second a million minutes of video content will cross the network. Yet, indexing, organizing, and even browsing such massive video data is still very challenging. 

As an attempt to mitigate the overload, \emph{video highlight detection} has attracted increasing attention in the research community.  %A closely related but different task is video summarization. 
The goal in highlight detection is to retrieve a moment---in the form of a short video clip---that captures a user's primary attention or interest within an unedited video.  \KG{A well-selected highlight can accelerate browsing many videos (since a user quickly previews the most important content), enhance social video sharing (since friends become encouraged to watch further), and facilitate video recommendation (since systems can relate unedited videos in a more focused way).}
Highlight detectors are typically \emph{domain-specific}~\cite{sun2014ranking,yao2016highlight,yang2015unsupervised,potapov2014category,panda2017weakly,liu2015thumbnail},
%others
meaning they are tailored to a category of video or keywords/tags like skiing, surfing, etc.  This accounts for the fact that the definition of what constitutes a highlight often depends on the domain, e.g., a barking dog might be of interest in a dog show video, but not in a surfing video.

\begin{figure}[t!]
\centering
\renewcommand{\tabcolsep}{0pt}
\includegraphics[width=1\columnwidth]{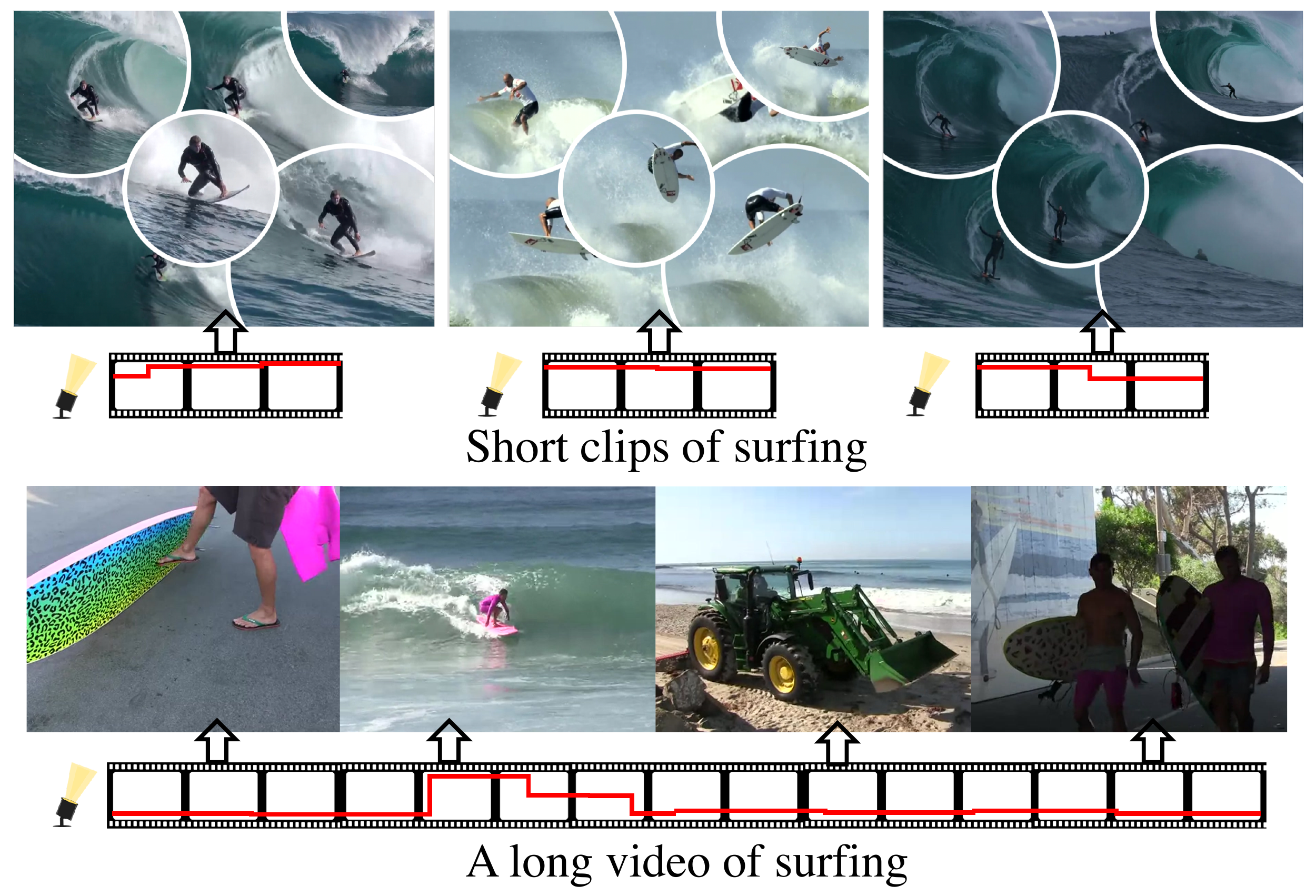}
\caption{ Video frames from three shorter user-generated video clips (top row) and one longer user-generated video (second row). Although all recordings capture the same event (surfing),  video segments from shorter user-generated videos are more likely to be highlights than those from longer videos, since users tend to be more selective about their content. The height of the red curve indicates highlight score over time. %\cc{Note most content from short clips has a high score whereas most content from the long video has low score.}  
\KGtwo{We leverage this natural phenomenon as a free latent supervision signal in large-scale Web video.}
}
\label{fig:intro}
\vspace{-10pt}
\end{figure}

\KG{Existing methods largely follow one of two strategies.  The first strategy poses 
highlight detection as a supervised learning task~\cite{gygli2016video2gif,sun2014ranking,yao2016highlight}.  %  any-others-here-supervised?
Given unedited videos together with manual annotations for their highlights, a ranker is trained to score highlight segments higher than those elsewhere in the video~\cite{gygli2016video2gif,sun2014ranking,yao2016highlight}.
While the resulting detector has the advantage of good discriminative power, the approach suffers from heavy, non-scalable supervision requirements.  The second  strategy instead considers highlight learning as a weakly supervised recognition task.  
Given domain-specific videos, the system discovers what appears commonly among the training samples, and learns to detect such segments as highlights in novel videos for the same domain~\cite{yang2015unsupervised,potapov2014category,panda2017weakly,liu2015thumbnail}. 
%others?
While more scalable in supervision, this approach suffers from a lack of discriminative power.  Put simply, repetition across samples does not entail importance.  For example, while all dog show videos might contain moments showing the audience waiting in their seats, that does not make it a highlight.}

We introduce a novel framework for domain-specific highlight detection that addresses both these shortcomings.  Our key insight is that user-generated videos, such as those uploaded to Instagram or YouTube, carry a latent supervision signal relevant for highlight detection: their duration.  
\RV{We hypothesize shorter user-uploaded videos tend to have a key focal point as the user is more selective about the content, whereas longer ones may not have every second be as crisp or engaging.}
\KG{In the spirit of Twain's quote above, more effort is required to film only the significant moments, or else manually edit them out later.  Hence duration is an informative, though implicit, training signal about the value of the video content.} \BX{See Fig.\ref{fig:intro}.} We leverage duration as a new form of ``weak" supervision to train highlight detectors with unedited videos.  Unlike existing supervised methods, our training data requirements are scalable, relying only on tagged video samples from the Web.  Unlike existing weakly supervised methods, our approach can be trained discriminatively to isolate highlights from non-highlight time segments.

Given a category (domain) name, we first query  Instagram  to mine public videos which contain the given category name as hashtags.   We use a total of 10M Instagram videos.
Since the hashtag Instagram videos are very noisy, \KG{and since even longer videos will contain some highlights}, 
we propose a novel ranking model that is robust to label noise in the training data. In particular, our model introduces a latent variable to indicate whether each training \KGtwo{pair} is valid or noisy.  We model the latent variable with a neural network, and train it jointly with the ranking function for highlight detection.  On two public challenging benchmark datasets (TVSum~\cite{song2015tvsum} and YouTube Highlights~\cite{sun2014ranking}), we demonstrate our approach improves the state of the art for domain-specific unsupervised highlight detection.\footnote{\KG{Throughout, we use the term \emph{unsupervised} to indicate the method does not have access to any manually created summaries for training.  We use the term \emph{domain-specific} to mean that there is a domain/category of interest specified by keyword(s) like ``skiing", following~\cite{potapov2014category,yang2015unsupervised,panda2017weakly,liu2015thumbnail}.}}

Overall, we make the following 
% novel 
contributions:
\begin{itemize}
\item We propose a novel approach to unsupervised video highlight detection that leverages user-generated video duration as \YK{an implicit}
% ``free" 
training signal.
\vspace*{-0.05in}
\item We propose a novel video clip deep ranking framework that is robust to noisily labeled training data.
\vspace*{-0.05in}
\item We train on a large-scale dataset that is one to two orders of magnitude larger than existing ones, and show that the scale \KG{(coupled with the scalablility of our model)} is crucial to success.
\vspace*{-0.05in}
\item On two challenging public benchmarks, our method substantially improves the state of the art for unsupervised highlight detection, \KGtwo{e.g., improving the next best existing method by 22\%.}
\end{itemize}

\section{Related Work}\label{sec:related}

\paragraph{Video Highlight Detection}
Many prior approaches focus on
 highlight detection for sports video~\cite{rui2000automatically,xiong2005highlights,tang2011detecting,wang2004sports}. Recently, supervised video highlight detection has been proposed for Internet videos~\cite{sun2014ranking} and first-person videos~\cite{yao2016highlight}. These methods all require human annotated \KGtwo{$\langle$highlight, source video$\rangle$}  pairs for each specific domain.  The Video2GIF approach~\cite{gygli2016video2gif} learns from GIF-video pairs, \KG{which are also manually created}. All supervised highlight detection methods require human edited/labeled ranking pairs. In contrast, our method does not use manually labeled highlights. \KG{Our work offers a new way} to take advantage of freely available videos from the Internet.

Unsupervised video highlight detection methods do not require video  annotations to train.  They can be further divided into methods that are \KG{domain-agnostic or domain-specific.}
\KG{Whereas a domain-agnostic approach like motion strength~\cite{mendi2013sports} operates uniformly on any video},
%The second group of methods requires category label at video level to train model for highlight detection. 
domain-specific methods train on a collection of videos of the same topic.  They leverage concepts like visual co-occurrence~\cite{chu2015video}, category-aware reconstruction loss~\cite{zhao2014quasi,yang2015unsupervised}, or
collaborative sparse selection within a category~\cite{panda2017collaborative}. Another approach is first train video category classifiers, then detect highlights based on the classifier scores~\cite{potapov2014category} or spatial-temporal gradients from the classifier~\cite{panda2017weakly}.
\KG{Like the domain-specific methods, our approach also tailors highlights to the topic domain; we gather the relevant training videos per topic automatically using keyword search on the Web.  Unlike any existing methods, we leverage video duration as a weak supervision signal.}

\vspace*{-0.1in}
\paragraph{Video Summarization} 

\KG{Whereas highlight detection (our goal) aims to score individual video segments for their worthiness as highlights,} \emph{video summarization} aims to provide a complete synopsis of the whole video, often in the form of a structured output, e.g., 
a storyline graph~\cite{kim2014reconstructing,xiong2015storyline}, a sequence of selected keyframes~\cite{lee2012discovering} or clips~\cite{gygli2014creating,zhang2018retrospective}. 
Video summarization is often formalized as a structured subset selection problem considering \KG{not just importance} but also diversity~\cite{gong2014diverse,Lu_2013_CVPR} and coherency~\cite{Lu_2013_CVPR}.
Supervised summarization methods focus on learning a visual interestingness/importance score~\cite{lee2012discovering,gygli2014creating}, submodular mixtures of objectives~\cite{gygli2015video,xu2015gaze}, or temporal dependencies~\cite{zhang2016video,zhang2018retrospective}.
Unsupervised summarization methods often focus on
low-level visual cues to locate important segments. 
\KG{Recent unsupervised and semi-supervised methods use recurrent auto-encoders to enforce that the summary sequence should be able to generate a
sequence similar to the original video~\cite{yang2015unsupervised,mahasseni2017unsupervised,zhang2018retrospective}.} Many rely on Web image priors~\cite{khosla2013large,song2015tvsum,kim2014joint,kim2014reconstructing} or semantic Web video priors~\cite{cai2018weakly}. 
\KG{While we also leverage Web data, our idea about duration is novel.  }

\vspace*{-0.1in}
\paragraph{Learning with Noisy Labels:} Our work is also related to learning from noisy data, a topic of broad interest in machine learning~\cite{natarajan2015learning,liu2016pami}.
 The proportion SVM~\cite{yu2013propto} handles noisy data for training SVMs where a fraction of the labels per group are expected to be incorrect, with applications to activity recognition~\cite{lai2014video}. Various methods explore how to train neural networks with noisy data~\cite{sukhbaatar2014training,reed2014training,li2017learning}.
\bx{Recent work on attention-based Multiple Instance Learning (MIL) helps focus on reliable instances using a differentiable MIL pooling operation for bags of embeddings~\cite{ilse2018attention}.  Inspired by the attention-based MIL, we propose a novel attention-based loss to reliably identify valid samples from noisy training data, but unlike~\cite{ilse2018attention}, 1) we have ``bags" defined in the space of ranking constraints, 2) our attention is defined in the loss space, not in the feature space, 3) our model predicts scores at the instance level, not at ``bag'' level, and 4) our attention mechanism is extended with multiple heads to take into account a prior for the expected label noise level.
}

\section{Approach}\label{sec:approach}

We explore domain-specific highlight detection trained with unlabeled videos. 
We first describe how we automatically collect large-scale hashtag video data for a domain (Sec.~\ref{sec:dataCollection}). Then we present our novel framework for learning highlights aided by duration as a training signal (Sec.~\ref{sec:ranking}). \KG{The results will show the impact of our method to find highlights in standard public benchmarks (Sec.~\ref{sec:results}).}

\subsection{Large-scale Instagram Training Video}\label{sec:dataCollection}

First we describe our data collection process. \RV{We choose Instagram as our source to collect videos because it contains a large amount of public videos associated with hashtags.} In addition, \KG{because Instagram users tend to upload frequently via mobile for social sharing, there is a natural variety of duration and quality}---some short and eye-catching videos, others less focused.
%which agree with our assumption that video segments from shorter videos are more likely to be highlights. 
The duration of a video from Instagram can vary from \bx{less than a second} to 1 minute.

\begin{figure}[t!]
\centering
\renewcommand{\tabcolsep}{0pt}
\includegraphics[width=0.9\columnwidth]{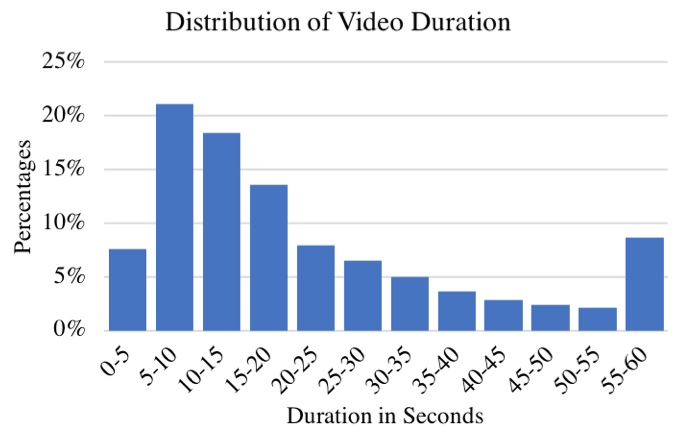}
\caption{Durations \KGtwo{for the 10M Instagram} training videos.
}
\label{fig:duration}
\end{figure}

Our goal is to build domain-specific highlight detectors. Given a category name, 
we query Instagram to mine for videos that contain the given category name among their hashtags. For most categories, this returns at least $200,000$ videos.  \KG{Since we validate our approach to detect highlights in the public TVSum and YouTube Highlights benchmarks~\cite{song2015tvsum,sun2014ranking} (see Sec.~\ref{sec:results}), the full list of hashtags queried are \emph{dog, gymnastics, parkour, skating, skiing, surfing, changing vehicle tire, getting vehicle unstuck, grooming an animal, making sandwich, parade, flash mob gathering, beekeeping, attempting bike tricks,} and \emph{dog show}.
Thus the data spans a range of domains frequently captured for sharing on social media or browsing for how-to's online.}  Altogether we acquire more than 10M training videos.  

\KGtwo{Figure~\ref{fig:duration} shows the distribution of their durations, which vary from less than a second to 1 minute.  We see there is a nice variety of lengths, with two modes centered around short ($\sim 10$ s) and ``long" ($\sim 60$ s) clips.}

\KG{Postprocessing hashtags, injecting word similarity models, or chaining to related keywords could further refine the quality of the domain-specific data~\cite{mahajan2018exploring}.  However, our experiments suggest that even our direct hashtag mining is sufficient to gather data relevant to the public video datasets we ultimately test on.  Below we will present a method to cope with the inherent noise in both the Instagram tags as well as the long/short video hypothesis.}

\subsection{Learning Highlights from Video Duration}\label{sec:ranking}

Next we introduce our ranking model that utilizes large-scale
hashtagged video data and their durations for training video highlight detectors.

Recall that a video highlight is a short video segment within a longer video that would capture a user's attention and interest.  Our goal is to learn a function $f(x)$ that infers the highlight score of a temporal video segment given its feature $x$ (\KGtwo{to be specified below}). Then, given a novel video, its highlights can be prioritized (ranked) based on each segment's predicted highlight score. 

A supervised regression solution would attempt to learn $f(x)$  from a video dataset with manually annotated highlight scores.  However, %\cc{this is problematic because} 
calibrating highlight scores collected from multiple human annotators %across a video dataset 
is itself challenging. \KGtwo{Instead,} highlight detection \KGtwo{can be} formalized as a \emph{ranking} problem by learning from human-labeled/edited video-highlight pairs~\cite{gygli2016video2gif,sun2014ranking,yao2016highlight}: \KG{segments in the \KGtwo{manually annotated} highlight ought to score more highly than those elsewhere in the original long video.} However, such paired data is difficult and expensive to collect, especially for long and unconstrained videos at a large scale.

\KG{To circumvent the heavy supervision entailed by collecting} video-highlight pairs, we propose a framework to learn highlight detection directly from a large collection of \KGtwo{\emph{unlabeled}} video.
As discussed above, we hypothesize that users tend to be more selective about the content in the shorter videos they upload, whereas their longer videos may be a mix of good and less interesting content.  
We therefore use the duration of videos as supervision signal. In particular, we propose to learn a scoring function that ranks video segments from shorter videos higher than video segments from longer videos. Since longer videos could also contain highlight moments, we devise the ranking model to effectively handle noisy ranking data.

\noindent{\bf{Training data and loss:}} Let $D$ denote a set of videos sharing a \KG{tag} (e.g., \emph{dog show}).  We first partition $D$ into three non-overlapping subsets $D=\{D_S,D_L,D_R\}$, where $D_S$ contains shorter videos, $D_L$ contains longer videos, and $D_R$ contains the rest.  \KGtwo{For example, shorter videos may be less than 15 seconds, longer ones more than 45 seconds} (cf.~Sec~\ref{sec:results}).
%we provide specifics of our partitioning in Sec~\ref{sec:results}.
\KG{Each video, whether long or short, is broken into uniform length temporal segments.\footnote{We simply break them up uniformly into 2-second segments, though automated temporal segmentation could also be employed~\cite{potapov2014category,song2015tvsum}.}}

Let $s_i$ refer to a unique video segment from the dataset, and let \KG{$v(s_i)$} denote the video where video segment $s_i$ comes from.  The visual feature extracted from segment $s_i$ is $x_i$.
%\KG{(e.g., we use I3D features~\ref{ref}). }
Since our goal is to rank video segments from shorter videos higher than those from longer videos, we construct training pairs $(s_i,s_j)$ such that $v(s_i)\in D_s$ and $v(s_j)\in D_L$.  We denote the collection of training pairs as \KG{$\mathcal{P}$}.  Since our dataset is large, we sample among all possible pairs, ensuring each video segment is included at least once in the training set. The learning objective consists of the following ranking loss:
\begin{equation}
\displaystyle
\begin{aligned}
 & L(D) =\sum_{(s_i,s_j)\in \mathcal{P}}\max\left(0,1-f(x_i)+f(x_j)\right),
\end{aligned}
\label{eq:ranking}
\end{equation}
\KGtwo{which says we incur a loss every time the longer video's segment scores higher.}
The function $f$ is a deep convolutional neural network, detailed below.  Note that whereas supervised highlight ranking methods~\cite{gygli2016video2gif,sun2014ranking,yao2016highlight} employ rank constraints on segments from the \emph{same} video---comparing those inside and outside of the true highlight region---our constraints span segments from distinct short and long videos.

\noindent\YK{{\bf{Learning from noisy pairs:}} The formulation \KGtwo{thus far} assumes that no noise exists and that $D_s$ and $D_L$ only contain segments from highlights and non-highlights, respectively.  However, this is not the case when learning from unedited videos:} some video segments from long videos can also be highlights, \KG{and some short segments need not be highlights.} 
\bx{Furthermore, some videos are irrelevant to the hashtags.}
\YK{Therefore, only a subset of our pairs in $\mathcal{P}$ have \textit{valid} ranking constraints $(s_i,s_j)$, i.e., pairs where $s_i$ corresponds to a highlight and $s_j$ corresponds to a non-highlight.} 
% ------------------------------------------------
Ideally, a ranking model would only learn from valid ranking constraints and ignore the rest. To achieve this without requiring any annotation effort, we introduce binary latent variables $w_{ij}$, $\forall (s_i,s_j) \in \mathcal{P}$ to indicate whether a ranking constraint is valid. We rewrite the learning objective as follows:
\begin{equation}
\displaystyle
\begin{aligned}
  & L(D) = \sum_{(s_i,s_j)\in \mathcal{P}} w_{ij}~~\max\left(0,1-f(x_i)+f(x_j)\right) \\
  & ~~~~~~~~~~~~~\textrm{s.t.} \quad \sum_{(s_i,s_j)\in \mathcal{P}} w_{ij}=p|\mathcal{P}|,\quad w_{ij}\in [0,1],\\
  & ~~~~~~~~~~~~~\textrm{and}~~w_{ij} = h(x_i,x_j)
\end{aligned}
\label{eq:EM}
\end{equation}
where $h$ is a neural network, $|\mathcal{P}|$ is total number of ranking constraints, and $p$ is the \KG{anticipated} proportion of ranking constraints that are valid.  \KG{In the spirit of learning with a proportional loss~\cite{yu2013propto}, this cap on the total weights assigned to the rank constraints represents a prior for the noise level expected in the labels. For example, training with $p=0.8$ tells the system that about 80\% of the pairs are a priori expected to be valid.}
The summation of the binary latent variable $w_{ij}$  prevents the trivial solution of assigning $0$ to all the latent variables.

% ------------------------------------------------
% Column-width Fig 2
% ------------------------------------------------
\begin{figure}[t!]
    \centering
    
    \includegraphics[width=\linewidth]{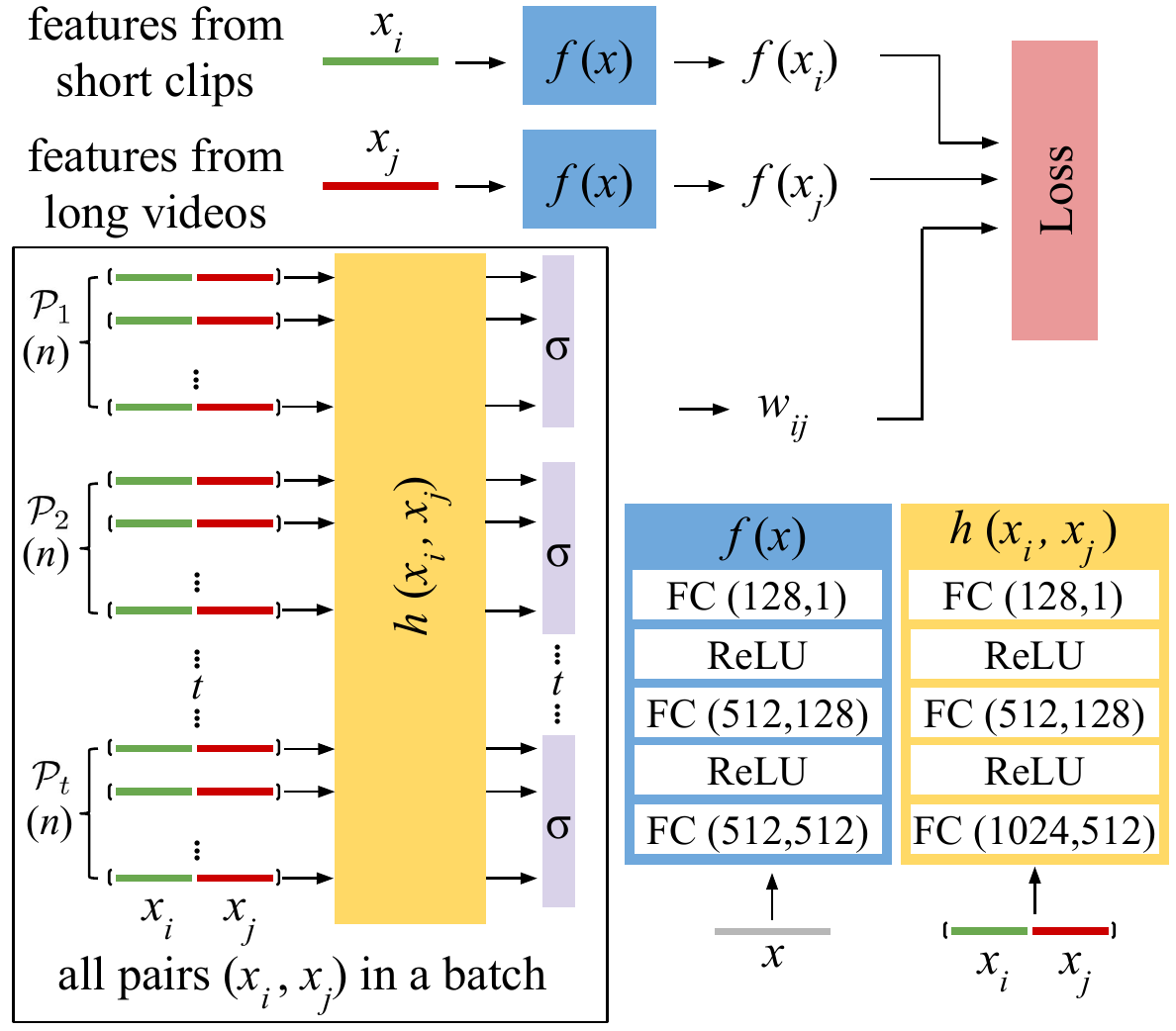}
    \caption{Network architecture details of our approach. \KGtwo{The} batch size is $b$. We group every $n$ instances of training pairs and feed them to a softmax function. Each batch has $t$ such groups ($b=nt$). 
    }
    \label{fig:app}
\end{figure}

\KG{Rather than optimize binary latent selection variables with alternating minimization, we instead use real-valued selection variables, and the function $h(x_i,x_i)$  directly predicts those latent variables $w_{ij}$.  The advantages are threefold. First,  we can simultaneously optimize the ranking function $f$ and the selected training pairs. Second, } 
the latent variable $w_{ij}$ is  conditioned on the input features so it can learn whether a ranking constraint is valid as a function of the specific visual input. Third, by relaxing $w_{ij}$ to a continuous variable in the range from $0$ to $1$, we capture uncertainty about pair validity during training.

\KGtwo{Finally, we parameterize the latent variables $w_{ij}$, which provide learned weights for the training \bx{samples}, and refine our objective to train over batches while enforcing the noise level prior $p$.}
\BX{We split the training data into groups, each of which contains exactly $n$ pairs. We then require that the latent variable $w_{ij}$ for instances within a group sum up to $1$.
In particular,} let $\mathcal{P}_1, \ldots, \mathcal{P}_m$ be a random split of the set of pairs $\mathcal{P}$ into $m$ groups where each group contains exactly $n$ pairs, then the final loss becomes:
\begin{equation}
\displaystyle
\begin{aligned}
  & L(D) = \sum_{g=1}^{m} ~ \sum_{(s_i,s_j)\in \mathcal{P}_g} \tilde{w}_{ij}~\max\left(0,1-f(x_i)+f(x_j)\right) \\
%   & ~~~~~~~~~~~~~\textrm{s.t.} \quad \sum_{(s_i,s_j)\in \mathcal{P}_g} \tilde{w}_{ij}=p|\mathcal{P}|,\quad \tilde{w}_{ij}\in [0,1], \\
  & ~~~~~~~~~~~ \textrm{s.t.} \sum_{(s_i,s_j)\in \mathcal{P}_g} \tilde{w}_{ij} = \sum_{(s_i,s_j)\in \mathcal{P}_g} \sigma_g(h(x_i,x_j)) = 1, \\
  & ~~~~~~~~~~~~~~ \tilde{w}_{ij}\in [0,1], 
\end{aligned}
\label{eq:attention}
\end{equation}
where $\sigma_g$ denotes the softmax function defined over the set of pairs in group $\mathcal{P}_g$.
\KGtwo{Note that now the group size $n$, together with the softmax, serves to uphold the label noise prior $p$, with $p=\frac{1}{n}$, while allowing a differentiable loss for the  selection function $h$.}
\KGtwo{Intuitively, smaller values of $n$ will speed up training at the cost of mistakenly promoting some invalid pairs, whereas larger values of $n$ will be more selective for valid pairs at the cost of slower training. In experiments, we fix $n$ to 8 for all results and datasets}. \bx{Please see Supp. for an ablation study with respect to $n$.}

\noindent {\bf Network structure:} We model both $f(x)$ and $h(x_i,x_j)$ with neural networks. We use a 
3 hidden layer fully-connected model for $f(x)$. The function $h(x_i,x_j)$ consists of a 3 fully-connected layers, followed by a $n$-way softmax function, \YK{as shown in Eq.(\ref{eq:attention})}. See Fig.~\ref{fig:app} for network architecture details.
% ------------------------------------------------

\noindent {\bf Video segment feature representation:} To generate features $x_i$ for a segment $s_i$ we use a 3D convolution network~\cite{hara3dcnns} with a ResNet-34~\cite{he2016deep} backbone pretrained on Kinetics~\cite{carreira2017quo}. We use the feature after the pooling of the final convolution layer. Each video segment is thus represented by a feature of 512 dimensions.  %\KG{\emph{Comment on motivation for this choice.}}

\noindent {\bf Implementation details:} We implement our model with PyTorch, and optimize with stochastic gradient with momentum for 30 epochs. \bx{We use a batch size of 2048 and set the base learning rate to 0.005.} We use a weight decay of 0.00005 and a momentum of 0.9.  \bx{ With a single Quadro GP100 gpu, the total  feature extraction time for a one-minute-long video  is 0.50 s.
After extracting video features, 
the total training time to train a model is one hour  for a dataset of 20,000 video clips of total duration 1600 hours}.  At test time, it takes 0.0003 s to detect highlights in a new one-minute-long  video after feature extraction.

\section{Results}\label{sec:results}

We validate our approach for highlight detection 
and compare to an array of \KGtwo{previous methods, focusing especially on those that are  unsupervised and domain-specific.}

\subsection{Experimental setup}

\noindent {\bf Datasets and metrics:} After training our model on the Instagram video, we evaluate it on two  challenging public video highlight detection datasets: YouTube Highlights~\cite{sun2014ranking} and TVSum~\cite{song2015tvsum}. YouTube Highlights~\cite{sun2014ranking} contains six domain-specific categories: \emph{surfing, skating, skiing, gymnastics, parkour}, and \emph{dog}. Each domain consists of around 100 videos and the total accumulated time is 1430
minutes.  TVSum~\cite{song2015tvsum} is collected from YouTube using 10 queries and  consists of 50 videos in total from domains including \emph{changing vehicle tire,  grooming an animal, making sandwich, parade,} \emph{flash mob gathering}, and others.   
Since the ground truth annotations in TVSum~\cite{song2015tvsum} provide frame-level importance scores, we first average the frame-level importance scores to obtain the shot-level scores, and then select the top 50\% shots \KG{(segments)} for each video as the human-created summary, following~\cite{panda2017collaborative,panda2017weakly}. Finally, the highlights selected by our method are compared with 20 human-created summaries. We report mean average precision (mAP) for both datasets.

\begin{table}\small
\setlength{\tabcolsep}{0.3em}
\hspace*{-0.15in}
\centering
\begin{tabular}{|c|c|c|c|c|c|c|c|}
\hline
            &  RRAE & GIFs  & LSVM & CLA & Ours-A & Ours-S \\ 
                       & \footnotesize{\KGtwo{(unsup)}~\cite{yang2015unsupervised}}& \footnotesize{\KGtwo{(sup)}~\cite{gygli2016video2gif}}  & \footnotesize{\KGtwo{(sup)}~\cite{sun2014ranking}} & \footnotesize{\KGtwo{(unsup)}}
                       & \footnotesize{\KGtwo{(unsup)}} &\footnotesize{\KGtwo{(unsup)}}  \\ \hline

dog        &   0.49  &   0.308          &  \textbf{0.60}  & 0.502 &  0.519              &   0.579               \\ \hline
gymnast.   &   0.35  &   0.335          &  0.41           & 0.217 &  \textbf{0.435}     &   0.417          \\ \hline
parkour    &   0.50  &   0.540          &  0.61           & 0.309 &  0.650              &   \textbf{0.670}     \\ \hline
skating    &   0.25  &   0.554          &  \textbf{0.62}  &  0.505    &  0.484               &   0.578    \\ \hline
skiing     &   0.22  &   0.328          &  0.36           &   0.379    &  0.410     &   \textbf{0.486}             \\ \hline
surfing    &   0.49  &   0.541          &  0.61           &  0.584&  0.531     &   \textbf{0.651}  \\ \hline\hline
Average    &   0.383 &   0.464          &  0.536          & 0.416 &  0.505     &   \textbf{0.564}             \\ \hline
\end{tabular}
\caption{Highlight detection results \KGtwo{(mAP)} on YouTube Highlights~\cite{sun2014ranking}. Our method outperforms all the baselines, including the supervised \KGtwo{ranking-based} methods~\cite{sun2014ranking,gygli2016video2gif}.}
\label{tab:youtube}
\end{table}

% \vspace{10pt}
\noindent\textbf{Baselines:} We compare with \KG{nine} state-of-the-art methods as reported in the literature. Here we organize them based on whether they require shot-level annotation (supervised) or not (unsupervised).  \KG{Recall that our method is unsupervised and domain-specific, since we \KGtwo{use no annotations} and compose the pool of training video with tag-based queries.}
\begin{itemize}
    \setlength{\itemsep}{-0.2\baselineskip}
\item \textbf{Unsupervised baselines:} %\cc{Unsupervised highlight detection methods do not require segment-level annotation.} 
We compare with the following unsupervised methods: RRAE~\cite{yang2015unsupervised}, MBF~\cite{chu2015video}, KVS~\cite{potapov2014category}, 
CVS~\cite{panda2017collaborative}, 
SG~\cite{mahasseni2017unsupervised}, 
DeSumNet(DSN)~\cite{panda2017weakly}, and VESD~\cite{cai2018weakly}.
\bx{We also implement a baseline where we train classifiers (CLA) with our hashtagged Instagram videos. The classifiers use the same network structures (except the last layer is replaced with a $K$-way classification) and video features as our method. We then use the classifier score for highlight detection. CLA \KGtwo{can be seen as a deep network variant of} KVS~\cite{potapov2014category}.
}

\item \textbf{Supervised baselines:} %\cc{Supervised video highlight detection methods require shot-level annotation.} 
We compare with the latent-SVM approach~\cite{sun2014ranking}, which trains with human-edited video-highlight pairs, and the Video2GIF approach~\cite{gygli2016video2gif}, a domain-agnostic method that trains with human-edited  video-GIF pairs. \KGtwo{Though these methods require annotations---and ours does not---they are of particular interest since they also use ranking-based formulations.}
\end{itemize}

\begin{table*}[t]\small
\centering
\begin{tabular}{|c|c|c|c|c|c|c|c|c|c|}
\hline
 & MBF~\cite{chu2015video} & KVS~\cite{potapov2014category} & CVS~\cite{panda2017collaborative} 
& SG~\cite{mahasseni2017unsupervised} 
& DSN~\cite{panda2017weakly}
& VESD~\cite{cai2018weakly}
& CLA
& Ours-A & Ours-S \\ \hline
Vehicle tire     & 0.295& 0.353 & 0.328 &0.423         &  -        &    -   &    0.294  &0.449             &  \textbf{0.559}\\ \hline
Vehicle unstuck  & 0.357& 0.441 & 0.413 &0.472         &  -        &    -   &    0.246  &\textbf{0.495}    &  0.429      \\ \hline
Grooming animal  & 0.325& 0.402 & 0.379 &0.475         &  -        &    -   &    0.590  &0.454             &  \textbf{0.612}   \\ \hline
Making sandwich  & 0.412& 0.417 & 0.398 &0.489         &  -        &    -   &    0.433  &0.537             &  \textbf{0.540} \\ \hline
Parkour          & 0.318& 0.382 & 0.354 &0.456         &  -        &    -   &    0.505  &0.602             &  \textbf{0.604} \\ \hline
Parade           & 0.334& 0.403 & 0.381 &0.473         &  -        &    -   &    0.491  &\textbf{0.530}    &  0.475 \\ \hline
Flash mob        & 0.365& 0.397 & 0.365 &\textbf{0.464}&  -        &    -   &    0.430  &0.384             &  0.432 \\ \hline
Beekeeping       & 0.313& 0.342 & 0.326 &0.417         &  -        &    -   &    0.517  &0.638             &  \textbf{0.663}\\ \hline
Bike tricks      & 0.365& 0.419 & 0.402 &0.483         &  -        &    -   &    0.578  &0.672             &  \textbf{0.691}\\ \hline
Dog show         & 0.357& 0.394 & 0.378 &0.466         &  -        &    -   &    0.382  &0.481             &  \textbf{0.626} \\ \hline\hline
\KGtwo{Average}  & 0.345& 0.398 & 0.372 &0.462         &0.424        &  0.423 &    0.447  &0.524    &  \textbf{0.563}     \\ \hline
%AP@15   &     &        &     &          &     &   &        \\ \hline
\end{tabular}
%\vspace{5pt}
\caption{Highlight detection results (Top-5 mAP score) on TVSum~\cite{song2015tvsum}. \KGtwo{All methods listed are unsupervised.}  Our method outperforms all the baselines by a large margin. \KGtwo{Entries with ``-" mean per-class results not available for that method.}} 
\label{tab:tvsum}
\end{table*}

We present results for two variants of our method: \textbf{Ours-A}: Our method trained with Instagram data in a domain-\emph{agnostic} way, where we pool training videos from all queried tags. We use a single model for all experiments; \textbf{Ours-S}: Our method trained with domain-\emph{specific} Instagram data, where we train a separate highlight detector for each queried tag.   \KG{For both variants, our method's training data pool is generated entirely automatically and uses no highlight annotations.  }
%, only the video durations as the training signal.}  
\KGtwo{A training video is in $D_S$ if its duration is between $8$ and $15$ s, and  it is in $D_L$ if its duration is between $45$ and $60$ s.}  \cc{We discard all other videos.} Performance is stable as long as we keep a large gap for the \KGtwo{two} cut off thresholds \KGtwo{(see Supp.)}.

\subsection{Highlight Detection Results}

\begin{figure*}[t!]
\centering

\includegraphics[width=2\columnwidth]{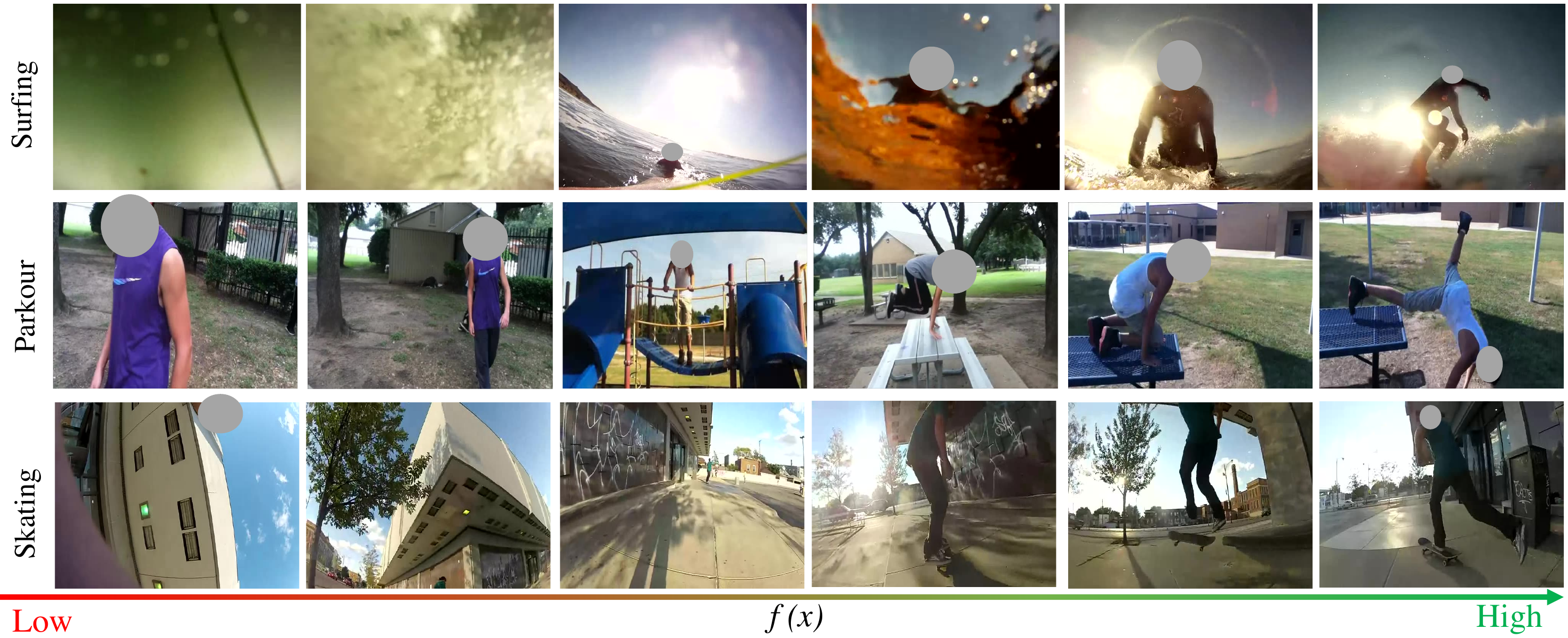}
\vspace{-10pt}
\caption{Example highlight detection results for the YouTube Highlights dataset~\cite{sun2014ranking}. We show our method's predicted ranking from low (left) to high (right) and present one frame for each video segment.  Please see Supp.~\KGtwo{video} for examples.}

\label{fig:qual}
\vspace{-10pt}
\end{figure*}

\vspace{5pt}
\noindent\textbf{Results on YouTube Highlights dataset:}
%\vspace{10pt}
Table ~\ref{tab:youtube} presents the results on YouTube Highlights~\cite{sun2014ranking}. 
\BX{All the baseline results are as reported in the authors' original papers.}
Our domain specific method (Ours-S) performs the best---\KGtwo{notably, it is even better than the \emph{supervised} ranking-based} methods. Compared to the unsupervised RRAE approach~\cite{yang2015unsupervised}, our average gain in mAP is 18.1\%.
Our method benefits from discriminative training to isolate highlights from non-highlight video segments.  
\bx{Our method also outperforms the CLA approach that is trained on the same dataset as ours, \KGtwo{indicating that our advantage is not due to the training data alone}. CLA can identify the most discriminative video segments, which may not always be highlights.}
\KGtwo{On average} our method outperforms the LSVM approach~\cite{sun2014ranking}, which is trained with domain-specific manually annotated data. 
\KGtwo{While the supervised methods are good at leveraging high quality training data, they are also limited by the practical difficulty of securing such data at scale.  In contrast, our method leverages large-scale tagged Web video at scale, without manual highlight examples.}

Our method trained with domain specific data (Ours-S) performs better than when it is trained in a domain-agnostic way (Ours-A). This is expected since highlights often depend on the domain of interest. 
\KGtwo{Still, our domain-agnostic variant} outperforms the domain-agnostic Video2GIF~\cite{gygli2016video2gif}, \KGtwo{again revealing the} 
benefit of large-scale weakly supervised video \KGtwo{for highlight learning}.

Fig.~\ref{fig:qual} \KGtwo{and the Supp.~video show example highlights.}  Despite not having explicit supervision, our method is able to detect highlight-worthy moments \KGtwo{for a range of video types.}

\vspace{5pt}
\noindent\textbf{Results on TVSum dataset:}
Table~\ref{tab:tvsum} presents the results on TVSum~\cite{song2015tvsum}.\footnote{\BX{The  results for CVS~\cite{panda2017collaborative}, DeSumNet~\cite{panda2017weakly} and VESD~\cite{cai2018weakly} are from the original papers. All others (MBF~\cite{chu2015video}, KVS~\cite{potapov2014category} and   SG~\cite{mahasseni2017unsupervised}) are as reported in~\cite{cai2018weakly}.}}
%Note VESD~\cite{cai2018weakly} reports results for both  an unsupervised and a supervised variant. We only compare with their unsupervised variant. }
\KGtwo{We focus the comparisons on unsupervised and domain-specific highlight methods.}
\cc{TVSum is a very challenging dataset with diverse videos. }
Our method outperforms all the baselines by a large margin. In particular, we outperform the next best method SG~\cite{mahasseni2017unsupervised} \bx{by 10.1 points, a relative gain of 22\%.} \BX{SG learns to minimize the
distance between original videos and their summaries.}
\KGtwo{The results reinforce the advantage of discriminatively selecting segments that are highlight worthy versus those that are simply representative.}
For example, while a close up of a bored dog might be more \emph{representative} in the feature space for dog show videos, a running dog is more likely to be a highlight. Our method trained with domain specific data (Ours-S) again outperforms our method trained in a domain-agnostic way (Ours-A). 

\vspace{5pt}
\noindent\textbf{\KGtwo{Instagram vs.~YouTube for training:}}
\KGtwo{Curious whether an existing large-scale collection of Web video might serve equally well as training data for our approach, we also trained our model on videos from YouTube8M~\cite{abu2016youtube}.  \bx{Training on 6,000 to 26,000 videos per domain from YouTube8M}, we found that results were inferior to those obtained with the Instagram data (see Supp.~for details).  We attribute this to two factors: 1) the YouTube-8M was explicitly curated to have fairly uniform-length ``longer" (120-500 s) clips~\cite{abu2016youtube}, which severely mutes our key duration signal, and 2) users sharing videos on Instagram may do so to share ``moments" with family and friends, whereas YouTube seems to attract a wider variety of purposes (e.g., instructional videos, edited films, etc.) which may also weaken the duration signal.}% We plan to make the Instagram URLs public to facilitate reproducibility.}

\subsection{Ablation Studies}

Next we present an ablation study to test variants of our method. 
%on YouTube highlight dataset. 
All the methods are trained with domain-specific data.  We compare our full method (Ours-S) with two variants:  1) \textbf{Ranking-D}, which treats all the ranking constraints as valid and trains the ranking function without the latent variables. \cc{This is similar to existing supervised highlight detection methods~\cite{gygli2016video2gif,yao2016highlight}.}
2) \textbf{Ranking-EM}, which introduces a binary latent variable and optimizes the ranking function and binary latent selection variable in an alternating manner with EM, similar to~\cite{sun2014ranking}. Note that \KGtwo{unlike our approach}, here the binary latent variable is not conditioned on the input and it takes discrete values.

 Table~\ref{tab:abaltion_method} shows the results. Our full method outperforms the alternative variants. \bx{In particular, our average gain in mAP over \textit{Ranking-D} is 13.9\% and 16.3\% for Youtube and TVSum respectively.} This \KGtwo{supports our hypothesis} that ranking constraints obtained by sampling training pairs $(s_i,s_j)$ such that $v(s_i)\in D_s$ and $v(s_j)\in D_L$ are indeed noisy. %; some video segments from long videos can  be highlights, and some short segments need not be highlights. 
 By modeling the noise and introducing the latent selection variable, our proposed method improves performance significantly. Our method also significantly outperforms \textit{Ranking-EM}, which also models noise in the training samples. In contrast to \textit{Ranking-EM}, our method directly predicts the latent selection variable from input. In addition, we benefit from joint optimization and relaxation of the latent selection variable, which accounts for uncertainty.%  during training. %\cc{\emph{SHOULD THIS also be shown for tvsum?}}

\begin{figure*}[t!]
\centering
\renewcommand{\tabcolsep}{0pt}
\includegraphics[width=2\columnwidth]{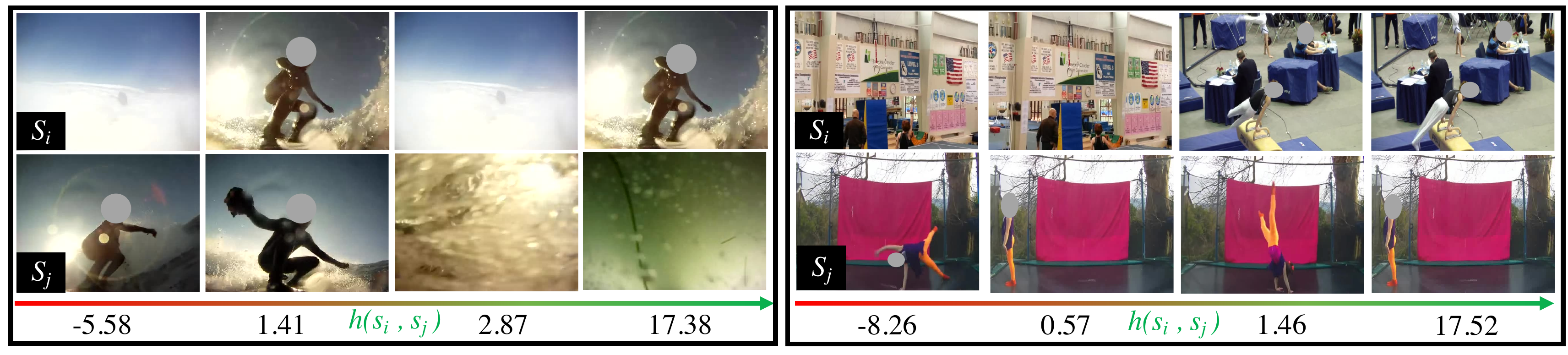}
\caption{ Predicted latent values (before softmax) for video segment pairs from YouTube Highlights. Higher  latent value indicates higher likelihood to be a valid pair. %\YK{Segments $s_i$ in top row come from short videos, segments $s_j$ in bottom row come from longer videos;} not true, all these are from YouTubes which we do not separate long from short.
The predicted latent value is high if $s_i$ (top row) is a highlight and $s_j$ (bottom row) is  a non-highlight. See Supp. for more.}

\label{fig:attention}
\vspace{-10pt}
\end{figure*}

\begin{table}[t]
\centering

\begin{tabular}{|c|c|c|c|}
\hline
Dataset    &  Ranking-D  & Ranking-EM &  Ours-S \\ \hline

YouTube    &   0.425    &   0.458          &  \textbf{0.564}              \\ \hline
TVSum     &   0.400    &   0.444          &  \textbf{0.563}              \\ \hline
\end{tabular}
%\vspace{5pt}
\caption{\KGtwo{Accuracy (mAP) in ablation study.} \bx{ Please see Supp. for per category results.}}
\label{tab:abaltion_method}
\vspace*{-0.1in}
\end{table}

\begin{figure}[t]
\centering
\renewcommand{\tabcolsep}{0pt}
\includegraphics[width=0.85\columnwidth]{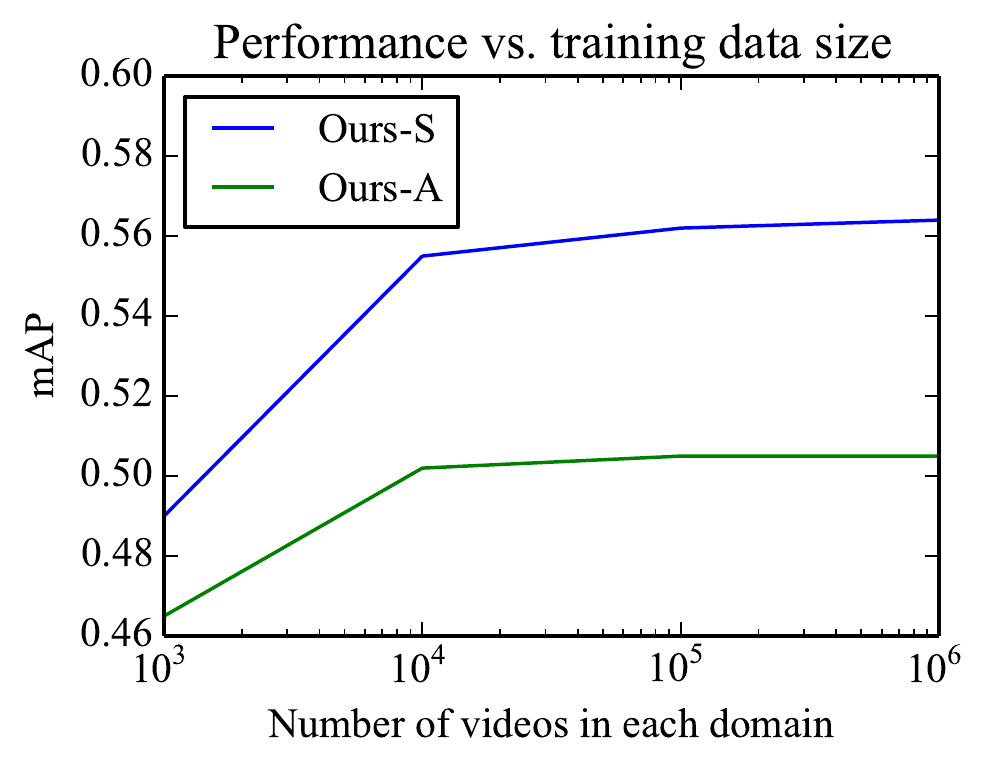}
\vspace*{-0.1in}
\caption{\KGtwo{Accuracy vs.~training set size} on YouTube~\cite{sun2014ranking}. %Performance improves as the training data size increases.
}
\label{fig:perform_num}
\vspace*{-0.15in}
\end{figure}

Fig.~\ref{fig:perform_num} shows highlight detection accuracy as a function of training set size.  \bx{We report this ablation for YouTube Highlights only, since the videos sharing tags with some TVSum categories max out at 24,000.} 
As we increase the number of videos in each domain, accuracy also improves. The performance improves significantly ($6.5\%$ for Ours-S and $3.7\%$ for Ours-A) when the training data is increased from $1,000$ to $10,000$ in each domain, \KGtwo{then starts to plateau}.

\subsection{Understanding Learning from Duration}\label{sec:understanding}

Finally, we investigate what each component of our model has learned from video duration. 
\KGtwo{First, we test whether our model can distinguish} segments from shorter videos versus  segments from longer videos.  \KGtwo{This is essentially a validation of the main training objective, without the additional layer of highlight accuracy.}
To answer the question, we train our model and reserve $20\%$ novel videos for testing. 
Each test pair consists of a randomly sampled video segment from a novel shorter video and one from a novel longer video.
We use $f(x)$ to score each \KGtwo{segment} and report the percentage of successfully ranked pairs. %(a video segment from shorter videos is ranked higher than a video segment from longer videos). 
\KGtwo{Without the proposed latent weight prediction}, our model achieves a $58.2\%$ successful ranking rate. 
Since it is higher than chance ($50\%$), this verifies our \KGtwo{hypothesis} that the distributions of the two video sources are different. However, the relatively low rate also indicates that the training data is very noisy.
After we weight the test video pairs with $h(x_i,x_j)$, we achieve a $87.2\%$ success rate.
The \KGtwo{accuracy} improves significantly because our latent value prediction function $h(x_i,x_j)$ identifies discriminative pairs.

\KGtwo{Second,} we \KGtwo{examine} video segment pairs constructed from the YouTube Highlights dataset \KGtwo{alongside} their predicted latent values (before softmax).  See Fig.~\ref{fig:attention}. Higher latent values indicate higher likelihood to be a valid pair. 
Video segments ($s_i$) from the top row are supposed to be ranked higher than video segments ($s_j$) from the second row. 
When $s_i$ corresponds to a highlight segment and $s_j$ a  non-highlight segment, the predicted latent value is high (last columns in each block). 
Conversely, the predicted latent value is extremely low when $s_i$ corresponds to a non-highlight segment and $s_j$ a highlight segment (first column in each block). Note if we group all the examples in each block into a softmax, all the training examples except the last will have negligible weights in the loss. 
This demonstrates that the learned $h(x_i,x_j)$ can indeed identify valid training pairs, and is essential to handle noise in training.

\section{Conclusions}\label{sec:Conclusions}
 
We introduce a scalable unsupervised solution that exploits \emph{video duration} as an implicit supervision signal for video highlight detection.  Through experiments on two challenging public video highlight detection benchmarks, our method substantially improves the state-of-the-art for unsupervised highlight detection.
\RV{The proposed framework has potential to build more intelligent systems for video preview, video sharing, and recommendations.}
  Future work will explore how to combine multiple pre-trained domain-specific highlight detectors for test videos in novel domains. 
\YK{Since the proposed method is robust to label noise and only requires weakly-labeled annotations like hashtags, it has the potential and flexibility to scale to an unprecedented number of domains, possibly utilizing predefined or learned taxonomies for reusing parts of the model.}

\onecolumn
 \section{Supplementary Material}\label{sec:sup}

 \subsection {Results on ablation study with respect to $n$ (Sec. 3.2 in main paper)}

This section accompanies Sec. 3.2 in the main paper.

\begin{figure*}[b!]
\centering
\includegraphics[width=0.5\columnwidth]{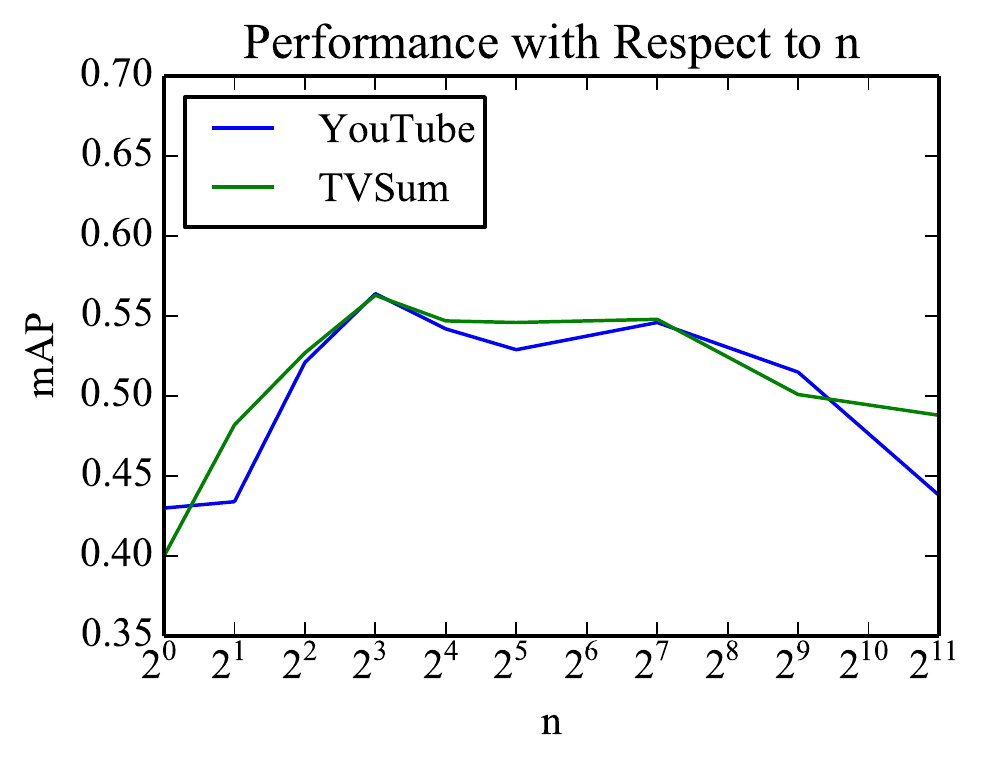}

\caption{Ablation study \KG{showing accuracy with respect to $n$.}} %We obtain the best performance when $n$ is 8. } 

\label{fig:num_n}
\end{figure*}

Recall that we split the training data into groups, each of which contains exactly $n$ pairs. \KG{Fig.~\ref{fig:num_n}} shows the results of \KG{an} ablation study with respect to $n$. We test $n$ from 1 to 2048 (the entire batch size). 
Note when $n$ is 1, the model is equivalent to the
\textit{Ranking-D} model because it assumes every training sample is valid. We obtain the best performance when $n$ is 8. Intuitively, smaller values of $n$ will speed up training at the cost of mistakenly promoting some invalid pairs, whereas larger values of n will be more selective for valid pairs at the cost of slower training. As $n$ becomes larger, the model only learns from a small proportion of valid pairs and ignores other valid ones, resulting in a drop in performance.

\subsection {Results on cut off thresholds for short and long videos (Sec. 4.1 in main paper)}

This section accompanies Sec. 4.1 in the main paper.

\begin{figure*}[t!]
    \centering
    \begin{subfigure}[t]{0.5\textwidth}
        \centering
        \includegraphics[width=1\columnwidth]{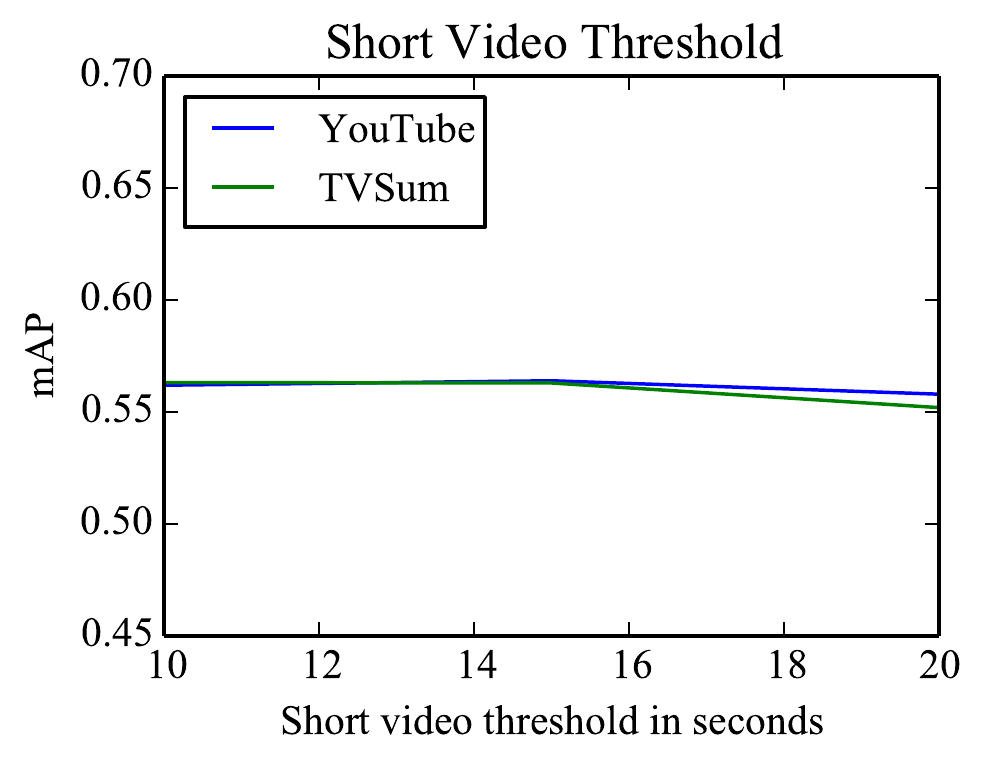}
        %\caption{Ablation study for cut off threshold for short videos. }
    \end{subfigure}%
    ~ 
    \begin{subfigure}[t]{0.5\textwidth}
        \centering
        \includegraphics[width=1\columnwidth]{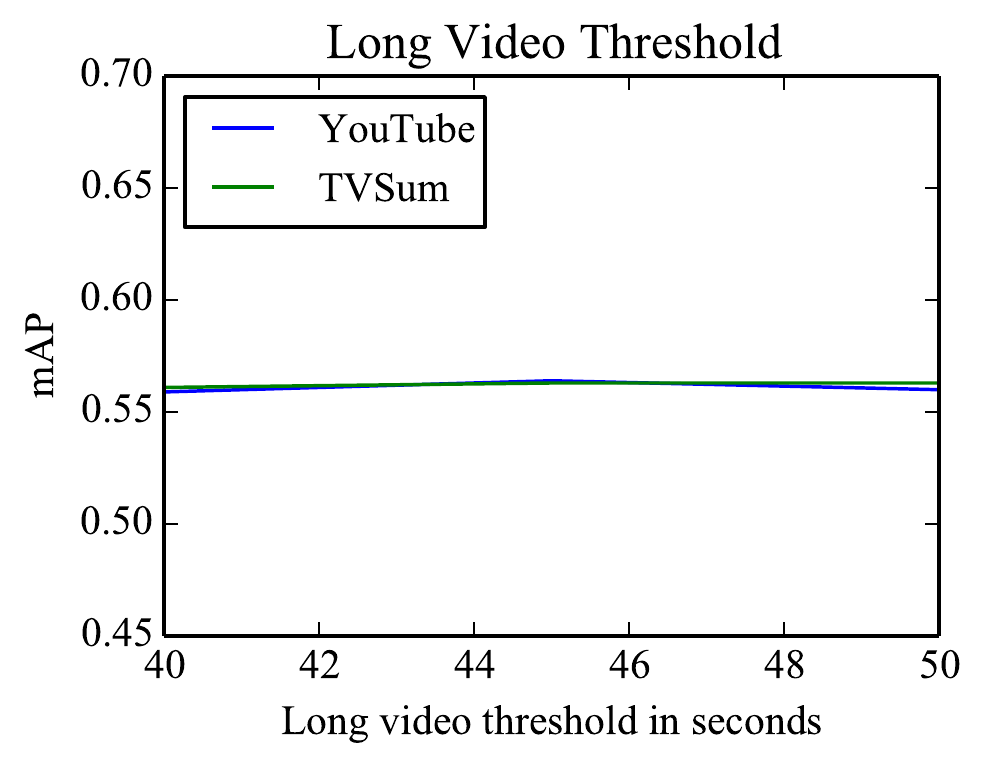}
        %\caption{Ablation study for cut off threshold for short videos. }
    \end{subfigure}
    \caption{Ablation study for cut off thresholds for short (left) and long (right) videos. \KG{Average accuracy (mAP)} on both datasets is insensitive to the cut off thresholds within the tested range.}
    \label{fig:supp_thred}
\end{figure*}

We show results on cut off thresholds for short and long videos in Fig.~\ref{fig:supp_thred}. On the left, we keep the threshold for long videos at 45 seconds and vary the threshold for short videos. On the right, we keep threshold for short videos at 15 seconds and vary the threshold for long videos. Performance on both datasets is insensitive to cut off thresholds within the tested range.

\subsection {Per category results for different variants of our methods (Sec. 4.3 in main paper)}
This section accompanies Sec. 4.3 in the main paper, \KG{where we summarized these results averaged over all categories.}

%We show per category results for an ablation study that tests different variants of our method.
Table~\ref{tab:tv_a} and Table~\ref{tab:youtube_a} \KG{show the per category results for both datasets}.  \KG{As also demonstrated in the main paper in Table 3}, our full method outperforms
the alternative variants for all video categories from both datasets. By modeling the noise and introducing the latent
selection variable, our proposed method improves performance
significantly.

\subsection {Highlight detection results using training data from YouTube 8M (Sec. 4.2 in main paper)}

\begin{figure*}[t!]
\centering

\includegraphics[width=0.5\columnwidth]{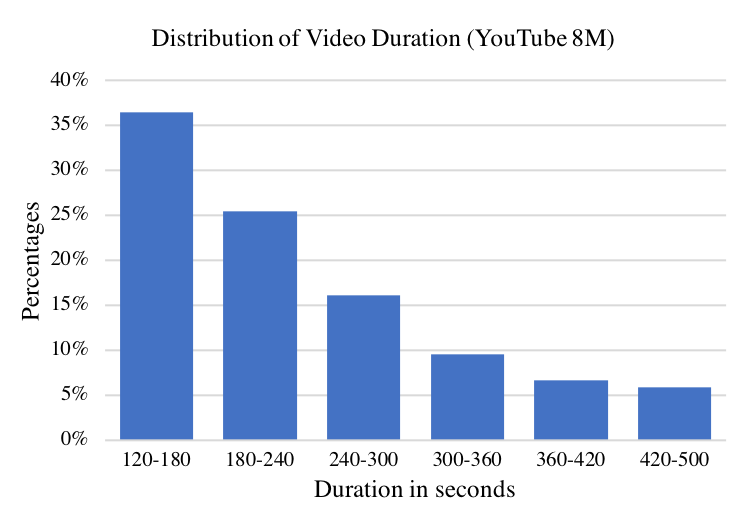}
\caption{Durations for the YouTube8M  training videos.}

\label{fig:supp_dura}
\end{figure*}

This section accompanies Sec. 4.2 in the main paper. 

We train our model with videos from YouTube8M~\cite{abu2016youtube}.   Fig.~\ref{fig:supp_dura} \KG{shows} the duration distribution.  \KG{We see that the distribution is quite different than that of the Instagram video collection, likely because of efforts made by the dataset creators to keep the videos on the longer side~\cite{abu2016youtube}.  In fact, the minimum length of any of these training videos is 120 seconds---longer than the ``long" videos in the Instagram data.  Thus this study seeks to understand the impact of the different sources for training.}

\KG{To train our model with this data, we let shorter videos be those} less than 180 seconds and longer ones be those more than 400 seconds (videos from YouTube8M range from 120 seconds to 500 seconds).
The video categories from YouTube8M~\cite{abu2016youtube} overlap with video domains from YouTube Highlights~\cite{sun2014ranking}, except for \emph{parkour}.

\begin{table}
\centering
\begin{tabular}{|c|c|c|c|c|c|c|c|c|c|}
\hline
            &  RRAE & GIFs  & LSVM & CLA & Ours-A & Ours-S & Ours-A & Ours-S \\ 
                       & \footnotesize{\KGtwo{(unsup)}~\cite{yang2015unsupervised}}& \footnotesize{\KGtwo{(sup)}~\cite{gygli2016video2gif}}  & \footnotesize{\KGtwo{(sup)}~\cite{sun2014ranking}} & \footnotesize{\KGtwo{(unsup)}}
                       & \footnotesize{\KGtwo{(YouTube,unsup)}}
                       & \footnotesize{\KGtwo{(YouTube,unsup)}}
                       & \footnotesize{\KGtwo{(Instagram,unsup)}} &\footnotesize{\KGtwo{(Instagram,unsup)}}  \\ \hline

dog        &   0.49  &   0.308          &  \textbf{0.60}  & 0.502 &0.304 &0.228&  0.519              &   0.579       \\ \hline
gymnast.   &   0.35  &   0.335          &  0.41           & 0.217 &0.284 &0.305&  \textbf{0.435}     &   0.417          \\ \hline
parkour    &   0.50  &   0.540          &  0.61           & 0.309 &0.286 &-&  0.650              &   \textbf{0.670}     \\ \hline
skating    &   0.25  &   0.554          &  \textbf{0.62}  & 0.505 &0.211 &0.210   &  0.484               &   0.578    \\ \hline
skiing     &   0.22  &   0.328          &  0.36           & 0.379 &0.340 &0.397   &  0.410     &   \textbf{0.486}             \\ \hline
surfing    &   0.49  &   0.541          &  0.61           & 0.584 &0.478 &0.644 &  0.531     &   \textbf{0.651}  \\ \hline\hline
Average    &   0.383 &   0.464          &  0.536          & 0.416 &0.317 &0.357* &  0.505     &   \textbf{0.564}             \\ \hline
\end{tabular}
\caption{Highlight detection results (mAP) on YouTube Highlights~\cite{sun2014ranking}. We show results of our model when trained with video from YouTube8M~\cite{abu2016youtube}.
Domain-specific result for parkour is unavailable since YouTube-8M does not contain parkour category.
* indicates average is computed over 5 categories.}
\label{tab:youtube}
\end{table}

Table~\ref{tab:youtube} shows the results.  
We show both domain-agnostic (Ours-A) and domain-specific (Ours-S) results \KG{(except for parkour because of data availability)}. The results of our model trained with YouTube8M \KG{are} inferior to those obtained with the Instagram data. We attribute this to two factors: 1) the YouTube-8M was explicitly curated to have ``longer" (120-500 s) clips~\cite{abu2016youtube}, which severely mutes our key duration signal (videos longer than 45 seconds are already considered as ``long'' in our Instagram data), and 2) users sharing videos on Instagram may do so to share ``moments" with family and friends, whereas YouTube seems to attract a wider variety of purposes (e.g., instructional videos, edited films, etc.) which may also weaken the duration signal.

The only exception is the surfing domain: our model trained with YouTube8M  is comparable to that trained with Instagram data and outperforms all other baselines. By visually inspecting surfing videos from YouTube8M, we found that ``short'' surfing videos  contain highlights \KG{more frequently} than ``long'' surfing videos.
%, whereas other video categories do not have this nice property. 

\begin{figure*}[t!]
\centering

\includegraphics[width=1\columnwidth]{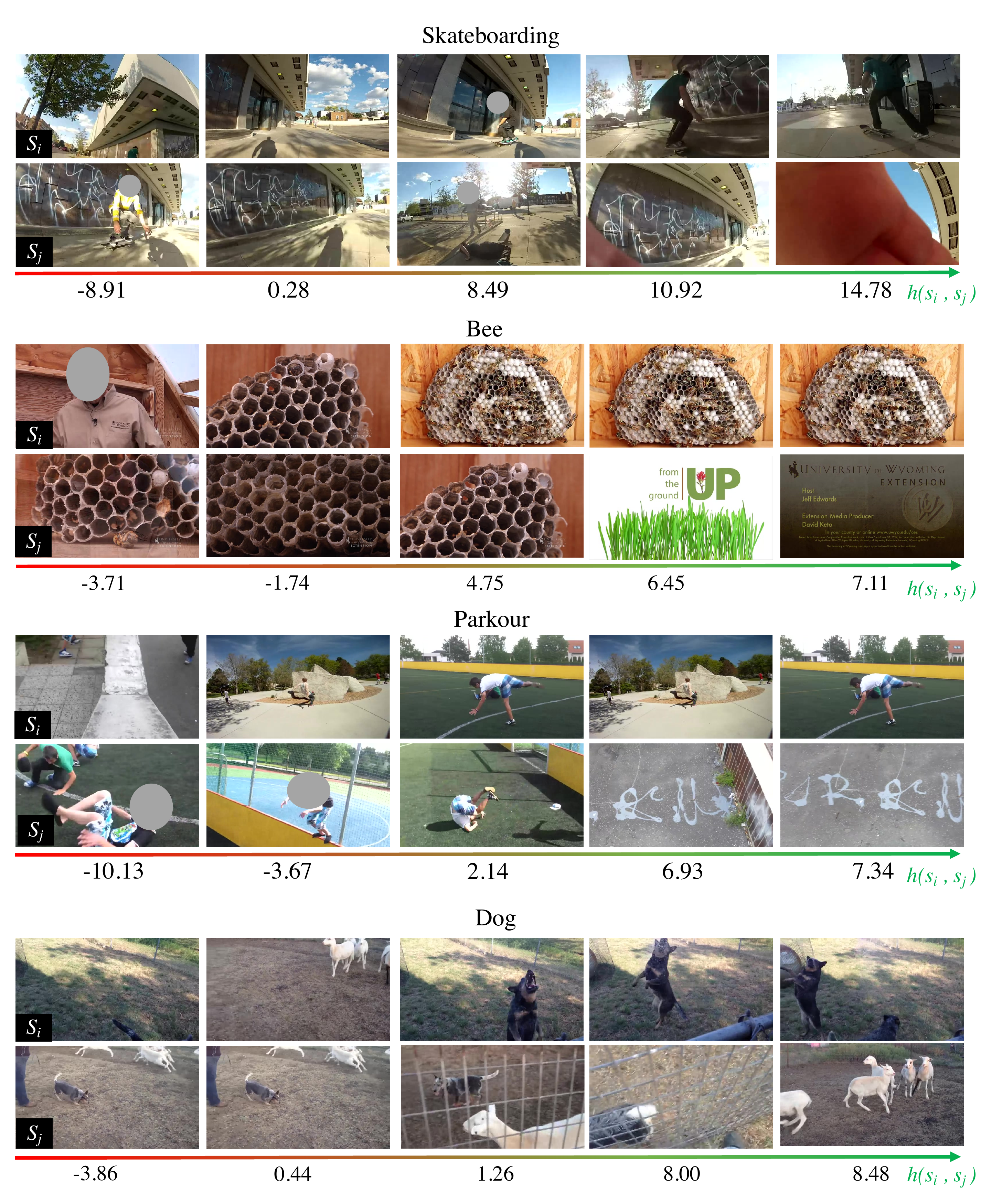}
\caption{
Predicted latent values (before softmax) for video segment pairs. Higher  latent value indicates higher likelihood to be a valid pair. 
The predicted latent value is high if $s_i$ (top row) is a highlight and $s_j$ (bottom row) is  a non-highlight.}

\label{fig:supp_attention}
\end{figure*}

\begin{table}[t]
        \begin{minipage}{0.5\textwidth}
            \centering
            \begin{tabular}{|c|c|c|c|}
\hline
    &  Ranking-D  & Ranking-EM &  Ours-S \\ \hline
dog        &   0.533    &   0.550          &  \textbf{0.579}                  \\ \hline
gymnastics &   0.282    &   0.304          &  \textbf{0.417}              \\ \hline
parkour    &   0.483    &   0.494          &  \textbf{0.670}        \\ \hline
skating    &   0.348    &   0.332          &  \textbf{0.578}        \\ \hline
skiing     &   0.460    &   0.482          &  \textbf{0.486}                \\ \hline
surfing    &   0.464    &   0.583          &  \textbf{0.651}                \\ \hline\hline
Average    &   0.425    &   0.458          &  \textbf{0.564}              \\ \hline

\end{tabular}
            \caption{\KG{Per domain} accuracy (mAP) in ablation study on TVSum~\cite{song2015tvsum}.}
            \label{tab:tv_a}
        \end{minipage}
        \hfill
        \begin{minipage}{0.5\textwidth}
            \centering
            \begin{tabular}{|c|c|c|c|c|c|c|c|c|c|}
\hline
 & Ranking-D & Ranking-EM & Ours-S \\ \hline
Vehicle tire     &    0.366   &  0.444           &  \textbf{0.559}\\ \hline
Vehicle unstuck  &    0.382   &  0.427          &  \textbf{0.429}      \\ \hline
Grooming animal  &    0.323   &  0.470          &  \textbf{0.612}   \\ \hline
Making sandwich  &    0.382   &  0.475          &  \textbf{0.540} \\ \hline
Parkour          &    0.460   &  0.472           &  \textbf{0.604} \\ \hline
Parade           &    0.395   &  0.372          &  \textbf{0.475} \\ \hline
Flash mob        &    0.333   &  0.277           &  \textbf{0.432} \\ \hline
Beekeeping       &    0.478   &  0.516          &  \textbf{0.663}\\ \hline
Bike tricks      &    0.574   &  0.561          &  \textbf{0.691}\\ \hline
Dog show         &    0.308   &  0.426          &  \textbf{0.626} \\ \hline\hline
Average  &    0.400   &0.444        &  \textbf{0.563}     \\ \hline

\end{tabular}
            \caption{\KG{Per domain} accuracy (mAP) in ablation study on YouTube Highlights dataset~\cite{sun2014ranking}.}
            \label{tab:youtube_a}
        \end{minipage}
    \end{table}

\subsection {More visualization examples on predicted latent values for video segment pairs (Sec. 4.4 in main paper)}
This section accompanies Sec. 4.4 in the main paper.  

We show more video segment pairs and their predicted latent values (before softmax).  See Fig.~\ref{fig:supp_attention}. Higher latent values indicate higher likelihood to be a valid pair. 
Video segments ($s_i$) from the top row are supposed to be ranked higher than video segments ($s_j$) from the bottom row. 
When $s_i$ corresponds to a highlight segment and $s_j$ a  non-highlight segment, the predicted latent value is high.
Conversely, the predicted latent value is extremely low when $s_i$ corresponds to a non-highlight segment and $s_j$ a highlight segment (first column in each example).

\clearpage
\twocolumn
{\small
\bibliographystyle{ieee}
\bibliography{egbib}
}

\end{document}